\documentclass[graybox, natbib, nosecnum]{svmult}
\bibpunct{(}{)}{;}{a}{}{,} 

\usepackage{mathptmx}       
\usepackage{helvet}         
\usepackage{courier}        
\usepackage{type1cm}        

\usepackage{makeidx}         
\usepackage{graphicx}        
\usepackage{multicol}        
\usepackage[bottom]{footmisc}
\usepackage[normalem]{ulem}	
\usepackage{hyperref}  
\usepackage{soul}   
\usepackage{amsmath,amssymb,amsfonts}
\usepackage{subcaption}

\DeclareMathAlphabet{\mathcal}{OMS}{cmsy}{m}{n}

\newcommand{\ignore}[1]{}

 \def\C{\mathcal{C}} 
\def\F{\mathcal{F}}  
 \def\I{\mathcal{I}} 
  \def\I{\mathcal{I}}

 \def\dN{\mathbb{N}} 
\def\dR{\mathbb{R}}

\begin{document}

\title*{Complexity of Planning}
\author{Kiril Solovey}
\institute{Kiril Solovey (\href{https://kirilsol.github.io/}{kirilsol.github.io}) \at Department of Aeronautics and Astronautics, Stanford University, CA, USA,  \email{kirilsol@stanford.edu}}
%
%
\maketitle
\section{Synonyms}
Computational complexity, time complexity. 

\section{Definitions}
Complexity analysis of complete algorithms for robot motion planning. 


\section{Overview}
This chapter is devoted to the study of complexity of complete (or exact) algorithms for robot motion planning. The term ``complete'' indicates that an approach is guaranteed to find the correct solution (a motion path or trajectory in our setting), or to report that none exists otherwise (in case that for instance, no feasible path exists). Complexity theory is a fundamental tool in computer science for analyzing the performance of algorithms, in terms of the amount of resources they require. (While complexity can express different quantities such as space and communication effort, our focus in this chapter is on time complexity.) Moreover, complexity theory helps to identify ``hard'' problems which require excessive amount of computation time to solve. In the context of motion planning, complexity theory can come in handy in various ways, some of which are illustrated here. 

First, when designing a motion planner, complexity analysis can predict the execution time of the algorithm as a function of the problem's input size. Importantly, this analysis is performed prior to the deployment of the algorithm in a real-world setting with physical robots. Additionally, it can be used to identify the most time-consuming components in a proposed method, which can then be replaced with more efficient subroutines, if available. Furthermore, complexity analysis can be used for qualitative comparison between a number of proposed methods, thus choosing the one with the lowest execution time. 

Secondly, from a more theoretical perspective, complexity theory can assist in identifying particularly challenging problems in motion planning. That is, for those ``hard'' problems (to be defined more precisely below), all algorithms are doomed to run for a tremendously long time for certain instances of the problem, even for seemingly easy scenarios, e.g., those involving a few obstacles and a simple robot representation. One such problem  is finding the shortest collision-free path for a rigid-body translating robot in a three-dimensional space amid static obstacles. In fact, many variants of motion planning are known to be hard. 

From a practical standpoint, the fact that a given problem is computationally hard can obviously be seen as bad news to the practitioner tackling it. However, this knowledge can suggest a few alternative lines of attack. First, there may be a ``relaxed'' variant of the problem that can solved efficiently, which meets practitioner's requirements. For instance, it may be the case that finding a short solution path, i.e., not necessarily the shortest possible, requires more modest amount of resources. Secondly, it may be useful to relax the completeness requirements of the algorithm and consider approaches that have milder theoretical guarantees. For instance, search-based methods \citep{CohETAL14} typically discretize the motion-planning problem into a grid according to a certain resolution. While such techniques by definition cannot be complete, they are widely used in practice due to their simplicity. Another popular method is sampling-based planners, which approximate the structure of the problem via random sampling. While the latter approach is incomplete as well, it often comes with the guarantee that a solution will be found eventually, i.e., when the number of samples is large enough. See more information on sampling-based planners in~\citep{HKS16}, and the chapters ``Sampling-Based Roadmap Planners'' and ``Sampling-Based Tree Planners''.

Lastly we wish to clarify that the hardness results do not necessarily imply that {all} the inputs to a given problem are hard,  only {some} of them. Consider for instance the problem of {integer linear programming} that is widely used in the world of engineering. It is known to be computationally hard, but in practice  many instances of the problem, including those involving  thousands of variables, can be solved rather quickly. 


\section{Key Research Findings}
This section is dedicated to basic aspects of complexity theory. 

\subsection{Elements of Computational Complexity}
\begin{figure}
    \centering
    \begin{subfigure}[b]{0.45\textwidth}
        \includegraphics[width=\textwidth,page=1]{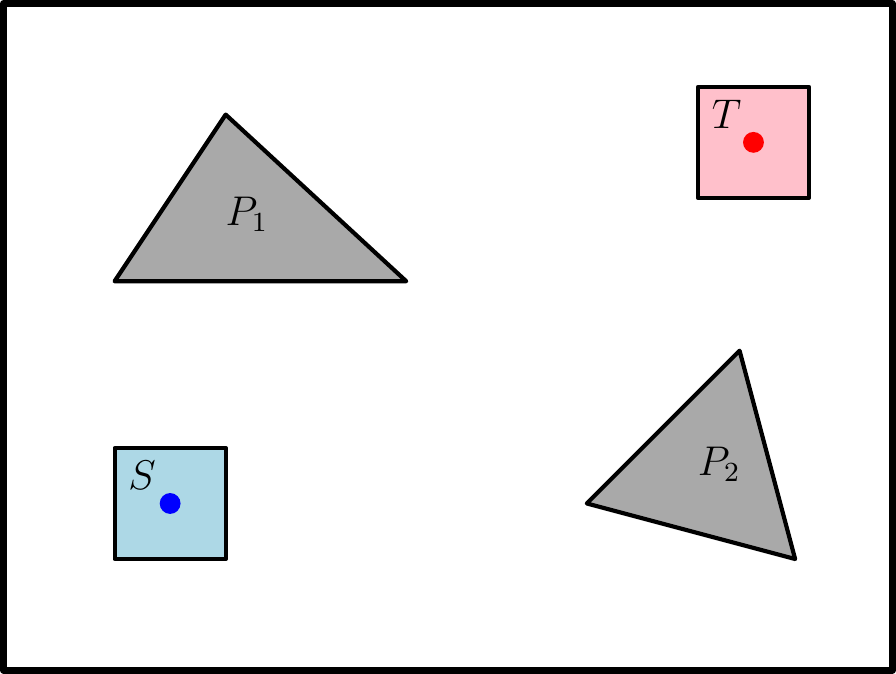}
        \caption{}
    \end{subfigure}
    ~ 
    \begin{subfigure}[b]{0.45\textwidth}
        \includegraphics[width=\textwidth,page=3]{configuration_space}
        \caption{}
    \end{subfigure}
    \caption{Motion planning for a square robot translating in the plane. In~(a) the start and target configurations $S,T$, are denoted by the blue and pink squares, respectively. The robot is confined to a room (black border) which contains two triangular obstacles (gray). The input size in this example consists of two variables, where $m=4$ denotes the complexity of the robot, and $n=3+3+4=10$ denotes the complexity (or the number of corners) of the obstacles.  In~(b), the forbidden regions of the configurations space are drawn in gray, whereas the white region is the free space. Observe that the depicted path is entirely collision free.}\label{fig:simple}
\end{figure}

The time complexity of an algorithm measures the amount of time required to achieve a solution for a given problem. {More generally, the complexity of a given problem (e.g., motion planing for a disc robot amid polygonal obstacles in the plane) refers to the time complexity necessary or sufficient to solve the problem.} As the notion of time highly depends on the system on which the algorithm is run, it is often more convenient and informative to measure the execution time by the amount of elementary operations (or basic steps) performed by the algorithm. Generally, it is desirable to understand how the size of the input for the problem affects the running time. The input size expresses the amount of information necessary to represent the input. For instance, for the problem of sorting an array of values, the size of input $n \in \dN$ would often represent the size of list. In the motion-planning problem depicted in Figure~\ref{fig:simple}, the input size  consists of two variables, which represent the complexity of the robot and workspace, respectively.

It is often difficult to nail down the precise running time of the algorithm with respect to the input size. Instead, an asymptotic expression capturing the performance for a sufficiently large input is typically derived. In particular, {big $O$ notation} is used to characterize the upper bound on the  running time according to its growth rate with respect to the input. The expression $O(f(n))$, where $f(n)$ is a function that depends on the size of the input $n$, indicates that there exists a constant $c$ such that for $n$ large enough the running time of the algorithm is at most $c\cdot f(n)$. ({The notation $\Omega$ similarly denotes the lower-bound on the running time, and $\Theta$ is used when the upper and lower bounds match.})
For instance, in the context of sorting an array of values, the {bubble sort} algorithm runs in time $O(n^2)$, whereas {merge sort} runs in time $O(n \log n)$. Thus, the latter algorithm is considered to be more efficient than the former. We mention that a complexity analysis of a given algorithm typically entails a meticulous study of its different ingredients. See~\cite{CormenETAL09} for more details. 

\subsection{Hardness of computation}
As hinted earlier, the majority of motion-planning problems are in fact computationally hard. Generally speaking, problems whose solution requires super-polynomial number of steps, e.g., complexity of $\Omega(2^n)$, are considered to be hard. Thus, it is important to establish whether a given problem admits a polynomial-time solution or not. For the brevity of exposition, we chose to omit most details concerning the classification of problems into easy and hard problems. We do mention that there exists a whole hierarchy between hard problems, and that the study of computational complexity is related to the famous problem of P vs NP.

{In the remainder of this chapter we will only distinguish between problems that are in the class of P ( polynomial complexity) and two classes devoted to hard problem, 
 namely NP-hard (or NP-complete) and PSPACE-hard (or PSPACE-complete), where the latter is considered to be more computationally demanding. The precise description of NP- and PSPACE-hardness requires several more definitions, and is beyond the scope of this chapter. For our purposes it would be sufficient to interpret problems that are NP-hard as requiring exponential running time (unless a widely accepted conjecture is false), regardless of the actual algorithm used. Problems that are PSPACE-hard require polyonimal space complexity and exponential time complexity. For a thorough treatment of this subject see~\cite{AroraBarak09}.}

Hardness proofs  are typically established via a technique of polynomial-time reduction. To prove that a given problem $A$ is computationally hard we rely on another problem $B$ that is already known to be hard. The approach requires to devise a reduction that  transforms in polynomial time any input of $B$ to an input to $A$, such that the input to $B$ is feasible (i.e., has a solution) if and only if the reduced input to $A$ is feasible. A polynomial-time reduction proves that if no efficient feasibility algorithm for $B$ exists, then none exists for $A$, as otherwise it would be possible to test the feasibility of $B$ by reducing it into $A$ and test the feasibility of the latter. 

There is a great variety of problems that are already known to be hard, and which can be used as our ``$B$'' problem for reduction purposes. A canonical example for a NP-hard problem is the 3SAT problem, which is concerned with finding a satisfying assignment to a Boolean expression consisting of binary variables $x_1,\ldots,x_n$, for some $n\in \dN_+$. The expression is of the form $C_1\wedge C_2\wedge \ldots \wedge C_k$, where $\wedge$ is an ``and'' operator, and  every $C_i$ represents a clause that consists of an ``or'' operator $\vee$ between three of the above variables or their negation. Consider for instance the following 3SAT expression with two clauses:
\[\phi:=(x_1\vee x_2\vee x_3)\wedge(\overline{x}_1\vee \overline{x}_2\vee x_3).\]
While it is easy to find a satisfying assignment for $\phi$, e.g., $x_1:=\text{true}, x_2:=\text{true}, x_3:=\text{true}$, in general, it is NP-hard to determine if a satisfying assignments exists when $n$, the number of variables, is no longer a constant. 

Many hardness proofs use reductions from 3SAT, including the hardness of shortest-path motion planning in 3D, that we mention below. While most such proofs are beyond the scope of this chapter, we provide a simple proof for the NP-hardness of integer linear programming (ILP) to illustrate the reduction technique. Recall that ILP consists of finding an integer assignment to a set of variables that satisfies a set of linear inequalities. Given an instance of a 3SAT problem, we transform it into ILP in the following manner. For every Boolean variable $x_i$ we assign an integer variable $z_i$ that can take values from $\{0,1\}$. We represent a clause $C_i$ as a linear inequality constraints which requires that the sum between the corresponding variables, or their negation, will be at least $1$. In particular, if a variable $x_j$ appears in its original form in the clause, we will include the variable $z_j$ in the sum, and otherwise we will include $(1-z_j)$. For example, the above 3SAT expression is reduced into the following two constraints:
\[z_1+z_2+z_3\geq 1,\quad (1-z_1)+(1-z_2)+z_3\geq 1.\]
It is straight forward to verify that this reduction applies to any 3SAT expression. In particular, a 3SAT expression has a satisfying assignment if and only if the reduced ILP problem has a solution. This proves that ILP is NP-hard as well. 

\section{Examples of Applications}
In this section we provide a more detailed account of complete approaches for motion planning of a single robot as well as for systems involving multiple robots. 

\subsection{Basic Motion Planning}
In its most basic form, motion planning consists of finding 
collision-free paths for a robot in a (two or three-dimensional)
{workspace} cluttered with static obstacles.  The spatial pose of
the robot, or its {configuration}, is uniquely defined by its
degrees of freedom (DoF). The set of all configurations $\C$ is
termed the {configuration space} of the robot, and decomposes
into the disjoint sets of free and forbidden configurations,
namely~$\F\subseteq \C$ and~$\C\setminus \F$, respectively. Thus, given
start and target configurations, the problem can be restated as the
task of finding a continuous curve in~$\F$ connecting the two
configurations. See Figure~\ref{fig:simple}.

The robot's number of DoFs, denoted by $d\in \dN_+$, is possibly the most crucial parameter in determining the complexity of the problem. Most complete algorithms for motion planning explicitly construct and maintain the robot's free space $\F$, and the dimension of this space is directly determined by $d$. Furthermore, the complexity incurred in representing $\F$ is typically exponential in $d$. For instance, the complexity of representing a single connected component of $\F$ can be as high as roughly $O\left(n^{d}\right)$~\citep{Basu03}. {It is therefore not surprising that motion planning in general is computationally hard when $d$ is part of the input. For instance, the problem is PSPACE-hard for a planar mechanical linkage robot with multiple links~\citep{HopcroftETAL84} and for for a multi-arm robot in a 3-dimensional polyhedral environment~\citep{Reif79}.}

{The first work to consider a complete method for a general dimension $d\geq 2$ leveraged cylindrical algebraic decomposition to design an approach with doubly-exponential running time on the order of $O\left({n}^{3^{d-1}}\right)$~\citep{ss-pm2}.  Later, a singly exponential algorithm termed the roadmap method was presented~\cite{Can93}. The latter has a running time on the order of $O\left({n}^{d^4}\right)$, which can be slightly lowered to $O\left({n}^{d^2}\right)$ if randomization is used. (See also a more detailed account of this method in the chapter ``Roadmaps''.)}


We proceed to consider specific cases of the problem with $d=2$, which admit an efficient solution. We start with the setting of a translating polygonal robot amid polygonal obstacles in the plane, as depicted in Figure~\ref{fig:simple}. In case the robot is convex, and the only obstacle in the environment is a convex room, then the problem can be solved in time $O(m+n)$, where $m$ is the complexity of the robot and $n$ is the complexity of the workspace~\citep{klps-ujr86}. In the more general case of non-convex robot and obstacles the problem can be solved in roughly $O(m^2n^2)$, where the complexity follows from computing $\F$ via Minkowski sums~\citep{AFH02}. We do mention that more refined results exist in the literature (see, e.g.,~\cite{HSS16}). 

{If we allow the polygon to translate and rotate, then the best known algorithm for this case runs in time $O\left((mn)^{2+\varepsilon}\right)$, for any $\varepsilon>0$~\citep{HS96}. However, restricting the robot's form to a rod or an L-shape lowers the complexity to $O(n^2)$~\citep{Veg90} or $O(n^2\log^2 n)$~\citep{HOS92}, respectively.}

{We now move to setting of a rigid body translating amid obstacles in $\dR^3$. The motion of a translating polyotope within a three-dimensional polyhedral environment can be computed in $O\left((mn)^{2+\varepsilon}\right)$ time~\citep{AS94}. In a similar setting of a ball robot in $\dR^3$, an algorithm with running time $O(n^{2+\varepsilon})$ exists, for any $\varepsilon>0$~\citep{AgarwalSharir00}. Although we are not aware of specialized approaches for general rigid-body robots translating and rotating in $\dR^3$ with completeness guarantees, there exists an $O(n^{4+\varepsilon})$-time algorithm for the special case of a rod robot~\citep{Koltun05}. A general search-based method for a rigid body translating and rotating in $\dR^3$ was presented in~\cite{Donald87}, and led to the development of multiple extensions and improvements in the literature. At its core, the method discretizes the continuous configuration space into a discrete grid of a certain user-defined resolution, and then applies a graph search method to retrieve a solution. By definition, this method is incomplete as the resolution necessary to obtain a solution is not known a priori. Moreover, for some problems in which the robot must touch obstacles to reach the target there exists no fixed resolution that allows the method to find a path.} 

\subsection{Optimal planning}
It is often desirable to obtain a high-quality solution path, or the best, i.e., {optimal}, solution possible. The quality can be associated with the length of the path traversed by the robot, time or amount of energy required to execute it, or safety that corresponds to the distance maintained between the robot and objects in its surrounding (e.g., obstacles or humans), to name just a few examples. 

{Typically, optimality constraints significantly increase the complexity of the problem, and in most cases these problems are computationally hard. For instance, for the 2D case, the shortest path for a point robot amid polygonal obstacles can be computed in $O(n\log n)$ time (this can be extended to a polygonal robot by first computing the free space). However, the three-dimensional extension for a translating robot is already NP-hard~\citep{Can88}, via a reduction from the 3SAT problem. (Also see additional information in chapter ``Roadmaps''.) We do mention that some special cases of the problem can be solved in polynomial time (see~\cite{Mitchell16}). For example, the shortest path for a point robot amid axis-parallel boxes in $\dR^3$ can be found in $O(n^2\log n)$ time~\citep{ChoiYap95}.}

{To alleviate the computational burden of the shortest-path problem for $d\geq 3$ approximation algorithms that relax the optimality guarantees have been developed. For instance, an $O\left(\frac{n^2}{\varepsilon^3}\log \frac{1}{\varepsilon}\log n\right)$-algorithm, which is guaranteed to return a solution of length at most $(1+\varepsilon)$, for any $\varepsilon\geq 0$, times the optimum, was presented in~\cite{LyuETAL10} for the setting of a point robot amid polyhedral obstacles in $\dR^3$. For additional information on approximation algorithms to this problem see~\cite{Mitchell16}.}

\subsection{Kinodynamic constraints}
So far our discussion has been mostly restricted to simple {geometric} robotic systems, in which the robot is assumed to be a rigid body free to translate or rotate as it pleases. However, in most real-life robotic systems the robot's motion is subject to kinodynamic or differential constraints, which must be satisfied in addition to collision avoidance. 

Addressing such constraints within a complete algorithmic framework turns out to be a challenging task. Consider for instance the seemingly simple case of a curvature-constrained planar model, also known as a {Dubins path}, which corresponds to a wheeled robot with a limited turning rate. The problem of finding the shortest such path turns out to be NP-hard~\citep{KKP11}, via a reduction from a generalized version of 3SAT. {Moreover, the same paper demonstrates that deciding (which is easier than finding a solution) whether there exists a non-intersecting Dubins path is NP-hard as well.} An approximation algorithm to this problem, which computes a solution whose length is at most $(1+\varepsilon)$ times the length of an optimal path, in $O\left(\frac{n^2}{\varepsilon^4} \log n\right )$ time, is given by~\cite{AgaWan00}.

{In the kinodynamic setting the path minimizing the travel distance (which we considered so far) can be different from the  minimal-time path minimizing the traversal time of the robot. For an intuitive example, consider a setting where a car needs to reach a target position location one kilometer away behind its back. The shortest path in this case will be simply backing up the car while it is in reverse gear. However, forward-driving gears are typically much faster, and so if the goal is to minimize time it will preferable to drive forward at an angle, thus forming a circle to reach the target with the car's front first.}

{Finding minimum-time path for a point robot in $\dR^3$  abiding particle dynamics amid polygonal obstacles is NP-hard~\citep{DonaldEA93}. The latter work also presents a polyonimal-time approximation algorithms for this problem which achieves an $(1+\varepsilon)$ approximation through a discretization of the state space. This approach was then extended to three dimensional multi-link robots with rigid-body dynamics~\citep{DonaldXavier95}, and then improved with respect to time complexity~\citep{ReifWang00}. Refer to chapter ``Kinodynamic Planning'' for more information on planning with constraints. }

\subsection{Multi-robot Planning}
In practical settings, such as in factory assembly lines, delivery services, and transportation systems, multiple robots are required to operate in a shared workspace, while avoiding collisions with each other, and at times cooperating to accomplish a mutual task. The entire fleet of robots can be viewed as one large robot having multiple moving parts. Thus, it is clear that the number of DoFs of the whole system increases with the number of robots. However, it should be noted that in multi-robot systems, the individual robots are rarely coupled or attached to each other. This property makes this problem slightly more manageable (as we will see later on), when compared to planning for a general system with the same number of DoFs.

One of the first algorithmic studies of multi-robot motion planning can
be found in the seminal series of papers on the {Piano Movers'
  Problem} by Schwartz and Sharir. They first considered the problem
in a general setting~(\cite{ss-pm2}) and then narrowed it down to the
case of disc robots moving amidst polygonal obstacles~(\cite{ss-pm3}).
In the latter work an algorithm was presented for the case of two and
three robots, with running time of $O(n^3)$ and $O(n^{13})$,
respectively, where $n$ is the complexity of the workspace.  Later~\cite{sy-snp84} used the {retraction method} to develop more
efficient algorithms, which run in $O(n^2)$ and $O(n^3)$ time for the
case of two and three robots, respectively.  Several years afterwards,
\cite{ss-cmp91} presented a general approach based
on {cell decomposition}, which is capable of dealing with additional types of robots and which has a running time of $O(n^2)$. 

When the number of robots is no longer a fixed constant the problem
can become computationally intractable. Specifically,~\cite{hss-cmpmio} showed that the problem is PSPACE-hard for the
setting of multiple rectangular robots bound to translate in a
rectangular workspace. \cite{sy-snp84} showed that
the problem is NP-hard for disc robots in a simple-polygon
workspace. 

{The results mentioned so far in this section deal with the labeled setting of the problem in which every robot is assigned to a specific target. In contrast, the unlabeled case the robots are given a
set of target positions and the goal is to move the robots in a
collision-free manner so that each robot ends up at {some}
target, without specifying exactly which one. Unfortunately, in general the latter problem remains computationally hard, as a recent paper~(\cite{SolHal16j}) presented a PSPACE-hardness
proof for {unlableled} unit-square robots translating amid
polygonal obstacles. However, several recent papers demonstrated that unlabeled motion planning for disc robots in the plane can be solved in polynomial time if one makes some simplifying {separation
  assumptions} with respect to the start and target configurations, as well as the locations of the obstacles in the workspace~\cite{tmk-cap13, abhs-unlabeled14, SolYuZamHal15}.}

\section{Future Direction for Research}
The practical applicability of most complete motion planners tailored for the single-robot case remains limited, even after 40 years of intensive research. For a few simple cases, polynomial-time algorithms exists, although it should be noted that most of them are practical only for small dimensions, e.g., $d\in \{2,3\}$, due to exponential dependence on $d$. Moreover, most problems are either NP-hard or PSPACE-hard, at the very least.  

One possible reason for existing approaches being so computationally expensive is their insistence on solving any feasible problem instance---including those that require the robot to get arbitrarily close to obstacles in order to reach its target. It has been already observed in sampling-based planning (see, e.g.,~\cite{TsaoETAL19}) that the difficulty of solving a motion-planning problem increases as its clearance decreases. The clearance of a problem $\delta\geq 0$ denotes the minimal distance of the robot from the obstacles that is necessary to be achieved in order to find a solution. In this context, most complete algorithms are designed to cope with an arbitrary value of $\delta$, even when it is equal to $0$. 

Thus, it would be interesting to study how the clearance of the problem affects its complexity. Alternatively, it may be worth to consider a relaxed family of algorithms that given a parameter $\delta>0$ are required to find a solution only if one exists with clearance at least $\delta$. {If this will turn out to be possible, the next step will be to extend those techniques to the more challenging setting of kinodynamic planning.}

\section{Cross-References}
Roadmaps; Sampling-Based Roadmap Planners (PRM and variations); Sampling-Based Tree Planners (RRT, EST and variations); Kinodynamic Planning.

\bibliographystyle{spbasic}  
\bibliography{bibliography} 


\end{document}